\documentclass[10pt, a4paper]{article}

\usepackage{lrec-coling2024} 

\usepackage{longtable}
\usepackage{rotating} 
\usepackage{tabularx} 
\usepackage{booktabs}

\usepackage{enumitem}

\usepackage{multirow}
\usepackage{multicol}
\usepackage{graphicx}
\usepackage{tipa}
\usepackage{amsmath}
\usepackage{afterpage}
\usepackage{cleveref} 
\crefname{table}{Tab.}{Tabs.} 
\Crefname{table}{Table}{Tables} 

\title{ART: The Alternating Reading Task Corpus for Speech Entrainment and Imitation}

\name{
    Zheng Yuan \ensuremath{\dagger}\thanks{\ensuremath{\dagger} Both authors contributed equally to this work.}$^{1,2}$,
    Dorina de Jong \ensuremath{\dagger}$^{1,2}$, 
    {\bf \large {\v{S}}tefan Be{\v{n}}u{\v{s}}$^{3,4}$}, 
    {\bf \large No{\"e}l Nguyen$^{5}$}, \\
    {\bf \large Ruitao Feng$^6$}, 
    {\bf \large R{\'o}bert Sabo$^4$},
    {\bf \large Luciano Fadiga$^{1,2}$}, 
    {\bf \large Alessandro D'Ausilio$^{1,2}$} \\
}

\address{
$^{1}$Italian Institute of Technology, Italy;
$^{2}$University of Ferrara, Italy \\
$^{3}$Constantine the Philosopher University in Nitra, Slovakia; 
$^{4}$Institute of Informatics, SAS, Slovakia \\
$^{5}$Aix Marseille University, CNRS, LPL, ILCB, France; 
$^{6}$Saarland University, Germany \\
\{zheng.yuan, dorina.dejong, luciano.fadiga, alessandro.dausilio\}@iit.it \\
sbenus@ukf.sk, robert.sabo@savba.sk, noel.nguyen-trong@univ-amu.fr \\ 
}

\abstract{
We introduce the Alternating Reading Task (ART) Corpus, a collection of dyadic sentence reading for studying the entrainment and imitation behaviour in speech communication. The ART corpus features three experimental conditions - solo reading, alternating reading, and deliberate imitation - as well as three sub-corpora encompassing French-, Italian-, and Slovak-accented English. This design allows systematic investigation of speech entrainment in a controlled and less-spontaneous setting. Alongside detailed transcriptions, it includes English proficiency scores, demographics, and in-experiment questionnaires for probing linguistic, personal and interpersonal influences on entrainment. Our presentation covers its design, collection, annotation processes, initial analysis, and future research prospects.
 \\ \newline  }

\begin{document}

\maketitleabstract

\section{Introduction}
Speech-based interpersonal communication is inherently dynamic and displays a number of interesting phenomena that suggest close synergetic coordination between interlocutors. Speech entrainment \citep{Levitan2011} is a phenomenon wherein the acoustic-prosodic characteristics of a speaker tend to become similar to those of their conversational partner. This observation is alternatively referred to as accommodation \citep{giles_accommodation_1991}, alignment \citep{Pickering2004}, convergence \citep{Pardo2006}, and imitation \citep{Goldinger1998} based on the research field and emphasis. 

Speech entrainment exerts diverse communicative effects, serving to foster rapport \citep{lubold2014acoustic}, facilitate collaborative tasks \citep{reitter2014alignment}, express identity \citep{soliz2014relational}, establish social distance \citep{earnshaw2021examining}, enhance language learning \citep{Lewandowski2019}, and potentially drive language change \citep{Gubian2023}. This phenomenon has been identified at multiple linguistic levels, spanning from lexicon \citep{brennan1996conceptual} to syntax \citep{reitter2010priming}, and is manifested through a range of acoustic-prosodic features \citep{Levitan2011}, including fundamental frequency \citep{Bradshaw2021} and vowel formants \citep{babel_evidence_2012}. It is evident in a range of conversational settings, including spontaneous dialogue \citep{cohen2020natural}, structured interactive tasks \citep{Pardo2006}, non-interactive tasks such as the shadowing task \citep{fowler_rapid_2003}, and even interactions between humans and computers\citep{coulston2002amplitude, bevnuvs2018prosodic}.


Speech entrainment is ubiquitous, yet its underlying mechanisms are notably intricate. Over the decades, researchers have embraced interdisciplinary perspectives and methodologies \citep{kruyt2023measuring} to unveil its nature and measure its degree, direction and dynamics during speech communication, encompassing social (e.g., the Communication Accommodation Theory, \citealp{giles_accommodation_1991, giles2016communication}), psycho-cognitive (e.g., the Interactive Alignment Model, \citealp{Pickering2004}; the Conversational Synergy Account, \citealp{fusaroli_dialog_2014}), and neuro-linguistic aspects (e.g., \citealp{ding2014cortical, mukherjee_neural_2019}). While research supports speech entrainment, studies show inconsistencies in their findings \citep{weise2019individual, pardo_special_2022, kruyt2023measuring}. This variability can be attributed to a multitude of factors influencing entrainment dynamics, ranging from individual speaker attributes (e.g., age, gender, personality, language and cultural background) to interactional variables (e.g., conversation role, social status) and experimental design (e.g., free or task-oriented interactions, audio-only or visual-audio settings). 

Another challenge is to explore the relationship between speech entrainment and imitation. Both processes likely share a foundation in similar brain regions for processing and producing speech \citep{delvaux_influence_2007, Garnier2013, sato2013converging}. While entrainment seems more subconscious mirroring another's speech patterns, imitation involves a more deliberate effort to copy specific sounds. Untangling how these processes interact could shed light on how our brains adapt and optimise language acquisition, particularly in second-language (L2) learning contexts.

To address the challenges, it becomes imperative to establish clear definitions of entrainment and its types according to a recognised framework (e.g., \citealt{wynn_classifying_2022}), meticulously design experiments to align with the research question at hand, and strive to control factors known to impact entrainment. Moreover, the experiment protocol shall include speech imitation data for comparison to ascertain whether entrainment occurs and to what extent it manifests in the specific interactions.

\citet{weise2022brooklyn} classified the commonly employed speech corpora in entrainment studies into two categories: those designed for general purposes and those specifically tailored for entrainment research. The former category includes the Switchboard Corpus \citep{godfrey1992switchboard}, the Fisher Corpus \citep{cieri2004fisher}, the CHAINS Corpus \citep{Cummins2006chain}, and the Columbia Games Corpus \citep{benus2007prosody}. The latter category comprises the Hcrc Map Task Corpus \citep{anderson1991hcrc} and its variations (e.g., \citealp{pardo2019montclair}), the Wildcat Corpus \cite{van_engen_wildcat_2010}, the SibLing Corpus \citep{kachkovskaia2020sibling}, and the Brooklyn Multi-Interaction Corpus \citep{weise2022brooklyn}, among others. 

Though valuable, existing resources primarily offer recordings of free conversation or word-level speech imitation. Since entrainment is "sparse" \citep{mukherjee2017relationship} and "subtle" \citep{weise2019individual}, these resources fall short of addressing our research question on acoustic-prosodic entrainment in L2-L2 interaction and its link to speech imitation. To bridge this gap, we developed the Alternating Reading Task (ART) Corpus. The next section describes its content, features, and design. For a comprehensive comparison of the listed corpora, readers are referred to  \ref{ss:comp}.


\section{The ART Corpus}
\label{sec:corpus}

The Alternating Reading Task (ART) Corpus \footnote{The initial release of the dataset is available for public access upon application (Italian \& French: https://zenodo.org/doi/10.5281/zenodo.4957145; Slovak: https://zenodo.org/doi/10.5281/zenodo.7993782).} was specifically designed to explore the acoustic-prosodic markers correlating speech entrainment with imitation, especially in L2-L2 interactions. Aligning with \citet{wilt2023automatic} that "automatic imitation is enhanced for non-native sound", we collected data from L2-L2 interactions. Its experiment settings can be generalised to any language pair and Human-Computer interactions (HCI), which enable potential applications in AI-powered language education, expressive speech synthesis and speaker recognition tasks. Key merits of this corpus include:


\begin{itemize}
\item The corpus includes solo, alternating, and imitation readings, facilitating comprehensive acoustic-prosodic entrainment and imitation studies.

\item Four rounds of alternating reading per dyad enable investigation of entrainment dynamicity (entrainment over time).
\item Italian-, French-, and Slovak-accented English sub-corpora with time-aligned transcripts support phonetic/phonological analysis in L2-L2 entrainment.

\item Four-dimensional spoken English scores (expert-evaluated) provide insights into language proficiency and entrainment.

\item Questionnaires assess partner perception (likeability, emotion, English proficiency, and imitation strategy), aiding analysis of social factors in entrainment.
\end{itemize}

\subsection{Experiment Design}
\label{design}

\begin{figure}[t]
    \centering
    \includegraphics[width=\columnwidth]{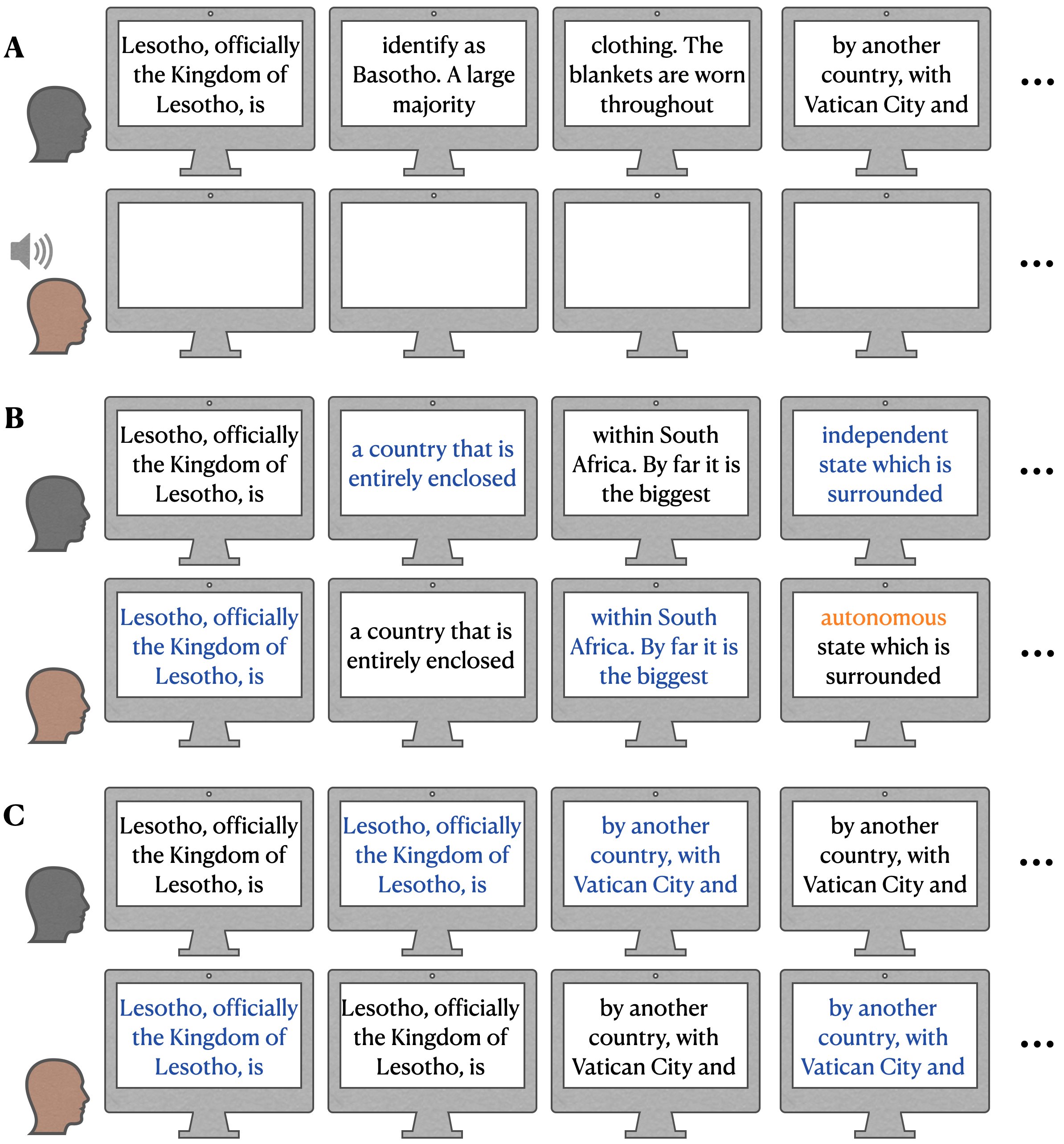}
    \caption{The Alternating Reading Task. Participants speak when their computer screen shows black sentences. (A) Solo Condition; (B) Interactive Condition - synonym shown in orange only for illustrative purpose, and (C) Imitation Condition.}
    \label{fig:task}
\end{figure}

The ART experiment is a collaborative speech production task that builds upon the Speech Domino paradigm \citep{bailly2010speech, bailly2014assessing, mukherjee2017relationship, mukherjee2018analyzing, aubanel2020speaking}. In this experiment, it extends the "domino" to the sentence/phrase level and introduces two additional experimental conditions for comparative analysis. These conditions are as follows:

\begin{description}
\item[Solo Condition] Participants read sentences from a neutral English text individually, serving as the baseline for individual voice characteristics.
\item[Interactive Condition] A pair of participants take turns reading aloud the text scripts over four rounds, with slight alterations (refer to \ref{proc} for details).
\item[Imitation Condition] The dyad is presented with a prompt to engage in mutual imitation where each person strives to speak in the same way as their partner.
\end{description}


This task design allows experimenters to control speech content while preserving a turn-taking structure similar to natural conversations. Furthermore, baseline and imitation conditions enable the assessment of implicit imitation (entrainment) and explicit speech imitation \citep{dufour2013much}.

\subsubsection{Material and Devices}

\begin{figure}[t]
    \centering
    \includegraphics[width=\columnwidth]{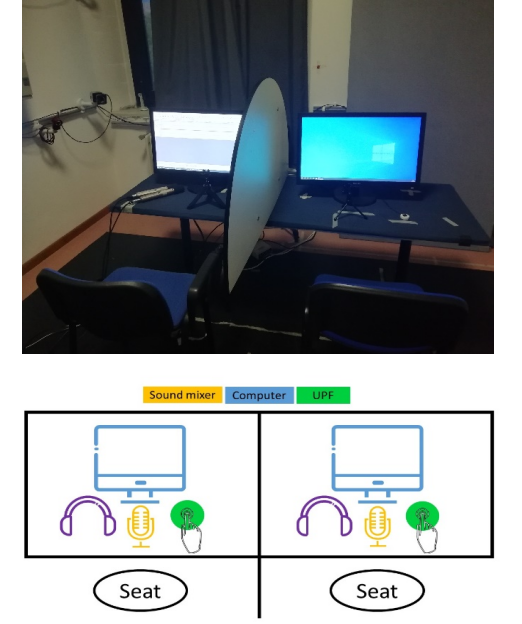}
    \caption{Picture and illustration of the ART setup with colour-coded connections: Microphones to sound mixer (orange), buttons to UPF box (green), computer screens to computer (blue). The sound mixer and UPF box are also connected to the computer, which runs necessary programs (PsychoPy \& Audacity)}
    \label{fig:setup}
\end{figure}

 The text employed in this experiment is a simplified adaptation of a Wikipedia article (\ref{Apd:art_text}) chosen for its emotional-neutral style. It consists of a total of 801 words and is segmented into 80 speaking turns, with word counts varying from 6 to 13 words. The strategic placement of turn boundaries within sentences was intended to promote prosodic continuity and mitigate long pauses between speaking turns \citep{aubanel2020speaking}. During the multiple rounds of alternating reading in the \texttt{interactive} condition, some words were replaced with their synonyms to maintain participants' attention and engagement (illustrated in Figure \ref{fig:task}B). 

Participants were seated side by side, facing two screens and separated by a curtain so that speech entrainment was not influenced by mutual visual contact \citep{schweitzer2017see}. The experiment was executed using Psychopy3 (v2021.1.2, \citealp{peirce2007psychopy}), and participants advanced to the next screen by activating a button connected to a UPF (USB to Parallel FIFO) interface \citep{canto2011convenient}. High-quality microphones (specifically AKG C1000S for the Italian and French experiments, and Sennheiser ME3 head-mounted close talk microphones for the Slovak experiment) were employed for speech recording via Audacity (Windows version 3.0.0 and 3.2), capturing audio at a 44.1 kHz sampling frequency. In our experiments, we used distinct sound mixers: a MAYA44 XTe audio interface for the Italian participants, an EDIROL UA-25 USB audio capture for the French participants, and a Focusrite Clarett 4Pre USB audio interface for the Slovak participants. 

\subsubsection{Participants}
A total of 58 participants were recruited for the ART experiment. This group comprised 18 native Italian (6 males, average age 24.50±3.65), 20 native French (all female, average age 23.45±4.94), and 20 native Slovak (10 males, average age 33.75±13.69). To ensure a minimum B2-level of English reading proficiency, all participants successfully passed the LexTALE \citep{lemhofer2012introducing} online test (test score: Italian=74.16±6.70, French=82.59±9.53, Slovak=78.12±10.24).
Participants were paired in same-sex dyads with similar LexTALE scores (<15\% difference in test scores)\footnote{Interlocutors' spoken English proficiency may have a larger difference according to the post hoc evaluation.}. The majority of participants were unacquainted with each other and had not communicated in English before the experiment. 

The demographic information, including age, sex, native language, and ethnic background, was collected as part of the informed consent process at the experiment's outset. It's essential to note that this sensitive data and the questionnaire responses were anonymised and not linked to participants' names or personal identifiers. All procedures complied with the Declaration of Helsinki and were approved by the local ethics committee.

\subsubsection{Procedure}
\label{proc}


Before the experiment, participants were required to complete a consent form. Following this, the experimenter provided participants with a detailed briefing on the experiment procedure and conducted device tests. The initial experimental condition, denoted as the \texttt{solo} condition (see Figure \ref{fig:task}A), involved participants reading aloud individually phrases displayed on the computer screen. These phrases constituted speaking turns from the alternating reading task, presented in a random order. While one participant was speaking, the other participant wore noise-cancelling headphones, listening to white noise to prevent unintended acoustic-prosodic entrainment to the speaker.

In the subsequent phase, participants performed alternating reading four times, with brief intermissions in between, constituting the \texttt{interactive} condition. We manipulated the presence of synonyms (see the words highlighted in orange in Figure \ref{fig:task}B) by replacing one word in a speaking turn with its synonym. Across these four rounds, the altered portion of turns commenced at 0\% and progressively increased to 75\%. Notably, the sequence of the four rounds for each dyad was determined randomly, but the speaker roles (e.g., who started the first turn) remained constant. The text appearing in white on the screen indicates the ongoing speaking turn, while the interlocutor views a (potentially differing) version of the text in grey on their own screen.

In the concluding phase, participants were explicitly directed to imitate each other. Phrases were presented randomly again for the initial speaker to recite. In the subsequent turn, the interlocutor encountered the same sentence and was instructed to replicate the speaker, without specific guidance on what aspects to imitate, such as pronunciation or intonation. This \texttt{imitation} condition (as depicted in Figure \ref{fig:task}C) served as both a control condition and a means to assess a participant's capacity for imitation.

Following each round of alternating reading, participants completed a reading comprehension test (\ref{apd_reading}) and a questionnaire (\ref{apd_questionaires}) assessing their perceptions of their interlocutor's friendliness, likability, social attractiveness, and level of relaxation. The reading comprehension tests serve as a refresher and a resource to explore how cognitive load affects speech entrainment. Slovak participants completed an additional questionnaire \footnote{The questionnaire was designed and added after the data collection for Italian and French speakers.} after the imitation condition, which inquired about their beliefs regarding their interlocutor's English proficiency and their attempts to adjust their pronunciation and intonation during the interactive condition, before receiving any specific instructions to imitate. All the questions were measured on a 5-point Likert scale. 

\subsection{English Proficiency Evaluation}
\label{English}

To investigate the potential impact of the L2 speaking proficiency of interlocutors on the degree of speech entrainment, a post hoc assessment of each speaker's proficiency in spoken English was carried out. Six language experts conducted this evaluation: three native Chinese speakers and three native Slovak speakers. The experts assessed the initial ten solo recordings of participants using four key criteria (detailed in \ref{apd_lang_criteria}): pronunciation, intonation, fluency, and overall impression. Each evaluator assigned scores on a scale ranging from 1 to 5 for each criterion. The final spoken English score for each speaker was computed as the average sum of these four indicators across the evaluators.



\begin{table}[t]
\centering
\begin{tabular}{@{}llll@{}}
\toprule[1pt] 
\textbf{Indicator} & \textbf{ICC} & \textbf{p-value} & \textbf{CI95\%} \\ 
\midrule
pronunciation & 0.828 & $<0.001$ & [0.71, 0.90] \\
intonation    & 0.767 & $<0.001$ & [0.65, 0.85] \\
fluency       & 0.796 & $<0.001$ & [0.68, 0.87] \\
overall       & 0.800 & $<0.001$ & [0.67, 0.88] \\
final         & 0.840 & $<0.001$ & [0.73, 0.91] \\
\bottomrule[1pt] 
\end{tabular}%
\label{tab:icc}
\caption{Intraclass Correlation Coefficients for Spoken English Proficiency Assessments}
\end{table}

Table \ref{tab:icc} illustrates the degree of agreement among these experts for each criterion, as quantified by the Intraclass Correlation Coefficients (ICC) values and their 95\% confidence intervals. The ICC values are based on a two-way mixed-effects model with a mean of raters and 57 degrees of freedom. All ICC values indicate a level of "good reliability" (ICC between 0.75 and 0.9) with statistical significance (p-value $<$ 0.001), aligning with the criteria stipulated by \cite{koo2016guideline}.

\subsection{Prepossessing and Transcription}
The recordings were automatically segmented into turn-level audio files relying on timestamps collected with Psychopy3 \citep{peirce2007psychopy}, yielding 18,560 segments with an average duration of 6.11s. Subsequently, the audio files underwent a review process to eliminate non-content segments, such as laughter and coughs, while preserving all instances of spoken words, including stutters, repetitions, and self-corrections. Stereo audio files were converted into mono audio files and were transcribed and force-aligned using the WhisperX ASR tool \citep{bain23_interspeech}. The transcriptions were then meticulously hand-calibrated to ensure accuracy and coherence\footnote{Due to the substantial size of the dataset, not all transcriptions received calibration in this initial release.}.

Mispronunciations in the recordings, typically influenced by the speaker's L1 background, are indicated in the transcription by appending the mispronounced form alongside the standard orthoepic form \footnote{This release includes a subset of annotated data.  Full annotations will be available in future versions.}. For example, some Slovak participants may articulate the long vowel \textipa{/i:/} as in "ski" closer to the diphthong \textipa{/aI/}, or they might pronounce the diphthong \textipa{/aI/} in "identify" closer to the short vowel \textipa{/i/}. These instances are transcribed as "ski (sk[\textipa{aI}])" and "identify ([\textipa{i}]dentify)". It is noteworthy that a subset of these mispronunciations self-corrected as the speaker progressed through alternating reading with a partner who consistently produced the standard pronunciation of the target words. This observation, coupled with the English proficiency data and psychological questionnaires, offers valuable insights for exploring acoustic-phonetic entrainment in the context of second language acquisition.

\subsection{Comparison with Other Corpora}
\label{ss:comp}


\begin{table*}
\centering
\begin{tabular}{p{4.2cm}p{1.8cm}p{1.8cm}p{1.8cm}p{1.8cm}p{2cm}}
\toprule[2pt]
\textbf{Corpus \newline Name} & \textbf{Session \newline Type} & \textbf{Language \newline Type} & \textbf{Speech \newline Baseline} & \textbf{Personal \newline Factor} & \textbf{Language \newline Proficiency} \\
\midrule
Switchboard \newline \citep{godfrey1992switchboard} & free & AME & No & No & No \\
Fisher \newline \citep{cieri2004fisher} & free & AME & No & No & No \\
CHAINS \newline \citep{Cummins2006chain} & free\& \newline task\& \newline shadowing & EHE & Yes & No & No \\
Columbia Games \newline \citep{benus2007prosody} & task & AME & No & No & No \\
Wildcat \newline \citep{van_engen_wildcat_2010} & task & AME, KOE, \newline ESE, TRE & No & No & Yes \\
SibLing \newline \citep{kachkovskaia2020sibling} & task & RU & No & No & No \\
Montclair Map Task \newline \citep{pardo2019montclair} & task\& \newline shadowing & AME & Yes & No & No \\
B-MIC \newline \citep{weise2022brooklyn} & free\& \newline task & AME & Yes & Yes & No \\
ART & task\& \newline shadowing & ITE, FRE, \newline SKE & Yes & Yes & Yes \\
\bottomrule [2pt]
\end{tabular}
\caption{Corpora Overview in Speech Entrainment Studies. Session Type indicates the register of recorded sessions, covering free-conversations, task-oriented interactions, and speech shadowing. Language Type abbreviations: AME for American English, EHE for  Eastern Hiberno-English, RU for Russian, KOE, ESE, TRE, ITE, FRE, and SKE for Korean-, Spanish-, Turkish-, Italian-, French-, and Slovak-accented English.}
\label{tab:corps}
\end{table*}


Table \ref{tab:corps} offers an overview of dialogue corpora that have been utilised in prior entrainment research, providing a comparative assessment alongside the ART Corpus in terms of their core attributes. The majority of subjects in these corpora engaged in task-oriented interactions, while the CHAINS Corpus \citep{Cummins2006chain} and B-MIC \citep{weise2022brooklyn} also feature free-form conversation. Notably, many of these corpora exhibit extended speaking turns, making the measurement of entrainment challenging due to the sporadic nature of natural entrainment \citep{mukherjee2017relationship}. In contrast, the Montclair Map Task provides word-level shadowing but lacks the capacity for investigating prosodic entrainment. The ART Corpus distinguishes itself by enabling a systematic exploration of speech entrainment across various speaking styles and proficiency levels, all within a controlled environment. Furthermore, it offers transcriptions with word alignment and psychological questionnaires related to the participants, which can contribute to the understanding of observed variations in entrainment behaviours across individuals, interlocutors, and communicative registers.

\section{Entrainment Experiments}
\label{sec:exp}

\afterpage{%
\begin{table*}[htp]
\centering
\begin{tabular}{lccccccccc}
\toprule[2pt]
\multirow{2}{*}{\textbf{Feature}} & \multicolumn{3}{c}{\textbf{IT Sub-Corpus}} & \multicolumn{3}{c}{\textbf{FR Sub-Corpus}} & \multicolumn{3}{c}{\textbf{SK Sub-Corpus}} \\
\cmidrule(r){2-4} \cmidrule(lr){5-7} \cmidrule(l){8-10}
& \textbf{SI} & \textbf{SM} & \textbf{MI} & \textbf{SI} & \textbf{SM} & \textbf{MI} & \textbf{SI} & \textbf{SM} & \textbf{MI} \\
\midrule
Mean Pitch & 15.90 & 16.09 & 11.48 & 18.31 & 18.61 & 13.68 & 12.48 & 12.08 & 8.51 \\
Max Pitch & 125.24 & 117.49 & 111.57 & 111.95 & 112.11 & 106.02 & 105.43 & 103.44 & 100.16 \\
Mean Intensity & 2.68 & 2.33 & 2.19 & 3.98 & 3.42 & 2.20 & 3.02 & 2.57 & 1.94 \\
Max Intensity & 3.02 & 2.59 & 2.36 & 3.97 & 3.42 & 2.31 & 3.68 & 3.34 & 2.69 \\
Jitter & 0.0034 & 0.0031 & 0.0029 & 0.0030 & 0.0029 & 0.0027 & 0.0047 & 0.0042 & 0.0035 \\
Shimmer & 0.0093 & 0.0093 & 0.0089 & 0.0091 & 0.0085 & 0.0076 & 0.0102 & 0.0097 & 0.0091 \\
HNR & 1.23 & 1.20 & 1.10 & 1.53 & 1.42 & 1.22 & 1.67 & 1.55 & 1.32 \\
Speech Rate & 0.575 & 0.538 & 0.433 & 0.504 & 0.506 & 0.435 & 0.481 & 0.484 & 0.409 \\
\bottomrule[2pt]
\end{tabular}
\caption{Inner-Speaker Distances for IT, FR, and SK Sub-Corpora}
\label{tab:in-spk}
\end{table*}

\begin{table*}[ht]
\centering
\begin{tabular}{lccccccccc}
\toprule[2pt]
\multirow{2}{*}{\textbf{Feature}} & \multicolumn{3}{c}{\textbf{IT Sub-Corpus}} & \multicolumn{3}{c}{\textbf{FR Sub-Corpus}} & \multicolumn{3}{c}{\textbf{SK Sub-Corpus}} \\
\cmidrule(r){2-4} \cmidrule(lr){5-7} \cmidrule(l){8-10}
& \textbf{SS} & \textbf{MM} & \textbf{II} & \textbf{SS} & \textbf{MM} & \textbf{II} & \textbf{SS} & \textbf{MM} & \textbf{II} \\
\midrule
Mean Pitch & 21.08 & 17.98 & 18.07 & 14.16 & 14.13 & 16.27 & 12.12 & 11.64 & 11.23 \\
Max Pitch & 136.66 & 132.69 & 129.59 & 130.83 & 139.81 & 146.2 & 127.87 & 126.78 & 124.18 \\
Mean Intensity & 7.05 & 7.40 & 7.59 & 6.38 & 5.02 & 6.10 & 2.49 & 2.18 & 2.13 \\
Max Intensity & 7.04 & 7.06 & 7.48 & 6.22 & 5.09 & 6.61 & 3.30 & 3.30 & 3.15 \\
Jitter & 0.0052 & 0.0052 & 0.0056 & 0.0039 & 0.0037 & 0.0043 & 0.0067 & 0.0048 & 0.0043 \\
Shimmer & 0.0134 & 0.0154 & 0.0154 & 0.0147 & 0.0143 & 0.0137 & 0.0175 & 0.0138 & 0.0131 \\
HNR & 1.75 & 2.09 & 2.08 & 3.07 & 2.66 & 2.88 & 2.81 & 2.5 & 2.23 \\
Speech Rate & 0.525 & 0.487 & 0.515 & 0.466 & 0.471 & 0.466 & 0.557 & 0.497 & 0.473 \\
\bottomrule[2pt]
\end{tabular}
\caption{Inner-Dyad Distances for IT, FR, and SK Sub-Corpora}
\label{tab:in-dyad}
\end{table*}

}

This section presents an initial analysis of the ART Corpus for speech entrainment. we investigated the global proximity \citep{Levitan2011, weise2022brooklyn, wynn_classifying_2022} at both inner-speaker and inner-dyad levels, across three experimental conditions: solo, interactive, and imitation. Global proximity is defined as the Euclidean distance between the speech feature values of two speakers over an entire session. We examined commonly used eight acoustic-prosodic features \citep{Levitan2011, weise2022brooklyn}. 

\subsection{Hypothesis}
Inner-speaker distance represents the absolute distance between the feature values of the same speaker across different experimental conditions. Conversely, inner-dyad distance signifies the absolute distance between the feature values of dyadic speakers within a singular experimental condition. Mathematically, let \( D(x) \) denote the distance between two sessions of the speaker(s). Specifically, \( D(\mathrm{SM}) \) is the inner-speaker distance between the solo and the interactive, or termed as, the main conditions, \( D(\mathrm{MI}) \) between the main and imitation conditions, and \( D(\mathrm{SI}) \) between the solo and imitation conditions. In terms of inner-dyad distances, \( D(\mathrm{SS}) \) represents the distance between the solo-solo conditions, \( D(\mathrm{MM}) \) between the main-main conditions, and \( D(\mathrm{II}) \) between the imitation-imitation conditions.

Our hypotheses for the study are as follows: 
\begin{itemize}
    \item \( H1 \): \( D(\mathrm{MI}) < D(\mathrm{SM}) < D(\mathrm{SI}) \)
    \item \( H2 \): \( D(\mathrm{II}) < D(\mathrm{MM}) < D(\mathrm{SS}) \)
    \item \( H3 \): \( D(\mathrm{MI}) < D(\mathrm{SM}) < D(\mathrm{SI}) < D(\mathrm{II}) \\
    < D(\mathrm{MM}) < D(\mathrm{SS}) \).
\end{itemize}

\subsection{Feature Extraction}
\label{subsec:feat}

The selected acoustic-prosodic features for our analysis include Mean Pitch (Hz), Max Pitch (Hz), Mean Intensity (dB), Max Intensity (dB), Jitter, Shimmer, Harmonics-to-Noise Ratio (HNR, in dB), and Speech Rate (syllables/second).

For the feature extraction, we utilised the Praat software (Windows version 6.3.19, \citealp{boersma2001praat}) along with the Parselmouth (version 0.4.3, \citealp{jadoul2018introducing}), a Python library that interfaces with Praat. All parameters during extraction were set to their default values.

\subsection{Processing Details}
\label{subsec:proc}
To compute the distance between features:
\begin{itemize}
    \item We first calculated the absolute distance between the same sentences in both the inner-speaker and inner-dyad settings. In the inner-speaker setting, distances were measured between two conditions, whereas in the inner-dyad setting, they were computed between the two dyadic speakers.
    \item The final distance for each feature was determined by taking the mean of the distances for all sentences, encompassing all speakers within a specific sub-corpus.
    \item In the solo-imitation (SI) setting, we took into account the direction of imitation. Only the sentences where the speaker imitated the partner were included.
    \item All outliers were included in our analysis without any exclusions.
\end{itemize}

\section{Results}
\label{sec:res}

In this section, we report the outcomes of the inner-speaker and inner-dyad distance experiments in speech entrainment. Tables \ref{tab:in-spk} and \ref{tab:in-dyad} offer a comprehensive portrayal of speech feature dynamics across different conditions and sub-corpora. A consistent pattern emerges, with inner-speaker distance increasing from solo to imitation conditions, while inner-dyad distance decreases over the same progression.

For hypothesis \( H1 \), 79.17\% of features across sub-corpora follow the expected inner-speaker distance trend, with 5 of 8 features adhering to \( H1 \). Max Pitch and Speech Rate align with \( H1 \) solely for the IT sub-corpus, and Max Pitch diverges for the FR sub-corpus. In inner-dyad analysis, \( H2 \) results reveal that 37.5\% of features exhibit the anticipated distance pattern, with the SK sub-corpus displaying the most consistency - 7 out of 8 features following \( H2 \). Max Intensity is the only feature deviating from \( H2 \) across all sub-corpora. IT sub-corpus follows \( H2 \) with Max Pitch, while the FR sub-corpus does so with Shimmer. The main condition that makes the results inconsistent is the imitation-imitation (\(\mathrm{II}\)) condition although the SK sub-corpus does follow \( H2 \). Consequently, \( H3 \) illustrates that 20.83\% of features conform to the combined trend of increasing inner-speaker distances and decreasing inner-dyad distances. Notably, Max Pitch and Shimmer consistently adhere to the hypothesised trends across all hypotheses.

\section{Discussions} 
\label{sec:disc}

The ART Corpus was designed to study the entrainment and imitation behaviours in L2-L2 speech communication. It enhances replicability in entrainment studies by offering a structured and less spontaneous interaction setting. Multiple short speaking turns with overlapping content are likely to induce entrainment more effectively than other forms of speech interaction, such as free conversation or map tasks. Furthermore, the varied experimental conditions, namely the solo, interactive, and imitation, offer an entrainment spectrum that enables direct comparison of entrainment degree. Researchers can investigate the dynamicity \citep{wynn_classifying_2022} of speech entrainment through multiple productions of the same text in progressive conditions.

Apart from recordings, other material, such as the spoken English score and the psychological questionnaires, open doors to linguistic and interpersonal factors of entrainment. Currently, the ART corpus provides Italian-, French-, and Slovak-accent English sub-corpora, however, from a wider perspective, its adaptable design can be employed across languages and conversation contexts. Thus, it holds potential applications in language education, speech technology, and therapeutic settings.

The global proximity experiments, despite complex hypotheses, demonstrate a consistent trend of entrainment progressing from the solo to the imitation conditions across various sub-corpora. These findings align with our previous research \citep{dejong22_interspeech, yuan23b_interspeech} on the ART Corpus using machine learning methods. We found that Max Pitch and Shimmer emerge as the most prominent features displaying entrainment patterns. Additionally, the degree of entrainment varies among sub-corpora. Admittedly, the valence of entrainment is possibly a combination of speech features \citep{weise2022towards} and the results are hardly comparable with other work featuring different experiment designs \citep{kruyt2023measuring}. Yet, our findings would be instrumental to future investigation of phonetic or prosodic entrainment using the ART corpus.


Looking forward, an immediate direction involves expanding the dataset size and diversity to improve the generalisability of findings. We should aim to include a broader range of speakers, encompassing more diverse linguistic backgrounds and proficiency levels. Adding subjective ratings of entrainment, e.g., perceived sentence similarity, would also be an important next step. As for entrainment studies, potential research directions include examining other entrainment types, for instance, local proximity at the inter-pausal unit (IPU) level, synchrony of pitch contour considering spoken English proficiency, and entrainment dynamics across alternating reading rounds. 







\section{Acknowledgements}
This work was partially supported by the European Union's Horizon 2020 research and innovation programme under the Marie Skłodowska-Curie grant agreement No 859588 and by the project COST CA19102 Language in the Human-Machine Era. We also extend our sincere thanks to Štefan Beňuš, Jana Beňušová, Lucia Mareková, Changyong Min, Qiuwen Zhang, and Fang Liu for their meticulous evaluation of the English language data.

\nocite{*}
\section{Bibliographical References}\label{sec:reference}

\bibliographystyle{lrec-coling2024-natbib}
\bibliography{Phonetic_Covergence}


\clearpage
\onecolumn
\appendix

\section{Appendices}

\subsection{Text Script for the ART Experiment}
\label{Apd:art_text}

\begin{longtable}{rl}
\toprule
 No. &                                                                                      Sentence \\
\midrule
\endfirsthead

\toprule
 No. &                                                                                      Sentence \\
\midrule
\endhead
\midrule
\multicolumn{2}{r}{{Continued on next page}} \\
\midrule
\endfoot

\bottomrule
\endlastfoot
    1 &                                 Lesotho, officially the Kingdom of Lesotho, is a country that \\
    2 &                                          is entirely enclosed within South Africa. By far, it \\
    3 &                                                       is the largest independent state, which \\
    4 &                                   is surrounded by another country, with Vatican City and San \\
    5 &                                   Marino being the other two. This makes Lesotho likewise the \\
    6 &   world's southernmost landlocked nation. The country is divided into ten districts, and each \\
    7 &                         of these districts is called after its principal towns. The country's \\
    8 &                                   capital and also its largest city is called Maseru. Lesotho \\
    9 &                                                          is the only independent state in the \\
   10 &                                     world that consists entirely above one thousand metres in \\
   11 &                                 elevation. Its lowest point of fourteen hundred metres is the \\
   12 &                      highest lowest point of any country in the world. Over eighty percent of \\
   13 &                                                          the country even lies above eighteen \\
   14 &                                               hundred metres. Thus, not surprisingly, Lesotho \\
   15 &                                    is likewise called the "Kingdom of the Sky." Likewise, you \\
   16 &                                    can find Africa's highest pub on the border of Lesotho and \\
   17 &            South Africa. Because of its elevation, Lesotho remains cooler throughout the year \\
   18 &                      than other regions at the same latitude. Snow is common in the highlands \\
   19 &                                    between May and September, and it is possible to go skiing \\
   20 &                                         on the slopes at that time. This makes Lesotho one of \\
   21 &                                                     the few places in Africa where you can go \\
   22 &                                                        skiing. Lesotho is home to the highest \\
   23 &                ski resort in Africa. Lesotho sees around three hundred days of sunshine every \\
   24 &                          year, and rainfall is highly variable because of its elevation. This \\
   25 &                                     can cause periodic droughts. Lesotho is mainly covered in \\
   26 &           grasses, although trees also appear on the landscape. Lesotho was formerly known as \\
   27 &             Basutoland, and almost the whole population of around two million people identify \\
   28 &                                      as Basotho. A large majority of the population practices \\
   29 &                                                  Christianity. Most families do their best to \\
   30 &                       be self-sufficient in food production, as food from South Africa can be \\
   31 &          very expensive. A staple food of the Basotho is cornmeal porridge. Particularly meat \\
   32 &                                                      and milk are rare for many households in \\
   33 &                           Lesotho, so cows are highly valued. Tea and locally brewed beer are \\
   34 &                                           popular beverages in the country. Lesotho's economy \\
   35 &                                                  is not surprisingly based on agriculture and \\
   36 & livestock, and approximately three-fourths of the population lives in rural areas. Mining for \\
   37 &   diamonds and manufacturing clothes are also activities that contribute significantly to the \\
   38 &                     economy of Lesotho. On another note, Lesotho is nearly self-sufficient in \\
   39 &                       electricity production, as the country generates a lot of hydroelectric \\
   40 &                                            power. The radio is the most popular form of media \\
   41 &                                                  in the country. Just a little bit over three \\
   42 &                                              percent of the population uses the Internet. The \\
   43 &                  official currency is the loti and can be used interchangeably with the South \\
   44 & African currency. Lesotho's official language is Sesotho. The name Lesotho roughly translates \\
   45 &                                    to "the land of the people who speak Sesotho." Sesotho was \\
   46 &                          one of the first African languages to develop a written form and has \\
   47 &                                          an extensive literature. Missionaries who arrived in \\
   48 &                                      Lesotho played a substantial role in this. Lesotho holds \\
   49 &                         one of Africa's highest adult literacy rates, with around eighty-five \\
   50 &                                                 percent for women and sixty-seven percent for \\
   51 &                          men. Lesotho is probably the only country in Africa where the female \\
   52 &                 literacy rate is much higher than the male literacy rate. High literacy rates \\
   53 &                                                     could result from primary education being \\
   54 &                                     free and compulsory for all children between ages six and \\
   55 &                                         thirteen. Football is the most widely played sport in \\
   56 &              Lesotho. Many of the country's most skilful players play professionally in South \\
   57 &                                    Africa. Horse racing is an important sport in rural social \\
   58 &                                                life. Most households in the rural areas own a \\
   59 &               small, sturdy Basotho pony for transportation and for helping out on the field, \\
   60 &                      along with donkeys. Lesotho's flag has three horizontal stripes in blue, \\
   61 &                       white, and green from top to bottom. The colours represent the motto of \\
   62 &                                   Lesotho: rain, peace, and prosperity. A traditional Basotho \\
   63 &                                                        hat is shown in black in the centre of \\
   64 &                                                     the flag. The title of Lesotho's national \\
   65 &                    anthem translates into "Lesotho, Land of Our Fathers." The Basotho blanket \\
   66 &                                                  is a thick colourful coat made primarily out \\
   67 &                                                  of wool. It is seen as an important piece of \\
   68 &           traditional attire. The blankets are worn throughout the country during all seasons \\
   69 &                      and worn differently by men and women. Although blanket styles have been \\
   70 &                                                 subject to outside influences, they are still \\
   71 &                                               closely linked with rites of passage in society \\
   72 &                  and certain Basotho's national events. Although modern Lesotho is only a bit \\
   73 &                                    older than 50 years, there are some interesting historical \\
   74 &                                  sites. For example, one of the largest dinosaur footmarks in \\
   75 &                                    the world has been discovered in Lesotho. Furthermore, you \\
   76 &                                           can find rock paintings of about one thousand years \\
   77 &                             old in remote caves. The Basotho people perform spiritual rituals \\
   78 &                           to treat illnesses and reduce misfortune in caves. Still, caves are \\
   79 &                    also places where rites of passages are being held. Caves are important to \\
   80 &                        the Basotho people, as they believe that their ancestors reside there. \\
\end{longtable}

\newpage
\subsection{In-Experiment Questionnaires}
\label{apd_questionaires}

Participants will fill out this form after each round of alternating reading.\\

\noindent \textbf{Participant number:} \\
\textbf{Round: } \\

(Here is the placeholder for the reading comprehension questions which have been relocated to Appendix A.3 for clarity and conciseness.) \\

Circle how strongly you agree with each statement

\renewcommand{\arraystretch}{1.5} 
\begin{longtable}{|p{0.35\textwidth}|p{0.1\textwidth}|p{0.1\textwidth}|p{0.1\textwidth}|p{0.1\textwidth}|p{0.1\textwidth}|}
\hline
\textbf{Statement} & \textbf{Strongly disagree} & \textbf{Disagree} & \textbf{Neutral} & \textbf{Agree} & \textbf{Strongly agree} \\ \hline
The other person is friendly & & & & & \\ \hline
The other person is likeable & & & & & \\ \hline
The other person is socially attractive & & & & & \\ \hline
The other person is relaxed & & & & & \\ \hline
\end{longtable}

The following is the additional questionnaire that the Slovak participants took after the imitation session.

\begin{longtable}{|p{0.35\textwidth}|p{0.1\textwidth}|p{0.1\textwidth}|p{0.1\textwidth}|p{0.1\textwidth}|p{0.1\textwidth}|}
\hline
\textbf{Statement} & \textbf{Strongly disagree} & \textbf{Disagree} & \textbf{Neutral} & \textbf{Agree} & \textbf{Strongly agree} \\ \hline
You think that the other person’s English is better than yours & & & & & \\ \hline
You have tried to adapt the pronunciation of English words to your partner in the INTERACTIVE session? (Even before you were asked for it in the imitation task) & & & & & \\ \hline
You have tried to adapt the intonation/melody of the sentences to your partner in the INTERACTIVE session? (Even before you were asked for it in the imitation task) & & & & & \\ \hline
\end{longtable}

\renewcommand{\arraystretch}{1}

\newpage
\subsection{Reading Comprehension Questions}
\label{apd_reading}

\begin{multicols}{2}

\subsubsection*{Round: 1}

Please answer the following questions:
\begin{enumerate}[leftmargin=*]
    \item What is the name of Lesotho’s capital?
    \begin{itemize}[leftmargin=*]
        \item Basotho
        \item Maseru
        \item Sesotho
    \end{itemize}
    
    \item Why are caves important to the Basotho people?
    \begin{itemize}[leftmargin=*]
        \item They believe their ancestors resided in caves
        \item They worship the rock paintings in the caves
        \item Caves keep them dry during the rainy season
    \end{itemize}
    
    \item What is NOT traditional Basotho clothing?
    \begin{itemize}[leftmargin=*]
        \item Basotho blanket/cloak
        \item Basotho hat
        \item Basotho shoes \\\\\\
    \end{itemize}
\end{enumerate}

\subsubsection*{Round: 2}

Please answer the following questions:

\begin{enumerate}[leftmargin=*]
    \item What is another name for Lesotho?
    \begin{itemize}[leftmargin=*]
        \item Kingdom of the Sky
        \item Kingdom of Mountains
        \item Kingdom of Mud
    \end{itemize}
    
    \item Which natural resource is mined in Lesotho?
    \begin{itemize}[leftmargin=*]
        \item Emerald
        \item Ruby
        \item Diamond
    \end{itemize}
    
    \item What is shown in the middle of the flag of Lesotho?
    \begin{itemize}[leftmargin=*]
        \item Bird
        \item Hat
        \item Cow
    \end{itemize}
\end{enumerate}

\subsubsection*{Round: 3}

Please answer the following questions:

\begin{enumerate}[leftmargin=*]
    \item Lesotho holds many records. What is NOT one of them?
    \begin{itemize}[leftmargin=*]
        \item Highest lowest point of any country in the world
        \item Highest pub/bar in the world
        \item Highest ski resort in Africa
    \end{itemize}
    
    \item What is the most widely played sport?
    \begin{itemize}[leftmargin=*]
        \item Football
        \item Horse racing
        \item Baseball
    \end{itemize}
    
    \item What is the motto of Lesotho?
    \begin{itemize}[leftmargin=*]
        \item Rain, peace and prosperity
        \item Rain, unity and equality
        \item Peace, unity and progress \\\\
    \end{itemize}
\end{enumerate} 

\subsubsection*{Round: 4}

Please answer the following questions:

\begin{enumerate}[leftmargin=*]
    \item What animal do many families own?
    \begin{itemize}[leftmargin=*]
        \item Cow
        \item Pig
        \item Horse
    \end{itemize}
    
    \item What is the most popular form of media?
    \begin{itemize}[leftmargin=*]
        \item Internet
        \item Radio
        \item Television
    \end{itemize}
    
    \item Most residents of Lesotho identify as:
    \begin{itemize}[leftmargin=*]
        \item Sesotho
        \item Mamotho
        \item Basotho
    \end{itemize}
\end{enumerate}
\end{multicols}

\newpage
\subsection{Language Proficiency Evaluation Criteria}
\label{apd_lang_criteria}

This is a guide to the evaluators.

\begin{enumerate}
    \item You will be presented with English sentences spoken by individuals with diverse first-language backgrounds.
    \item Listen attentively to each utterance and evaluate the speaker's English proficiency in terms of pronunciation, intonation, and fluency. Provide an overall rating considering these aspects.
    \item Refer to the detailed descriptions of speaking skill indicators and assign ratings on a scale from 1 to 5.
\end{enumerate}

\begin{longtable}{p{0.1\textwidth}p{0.25\textwidth}p{0.25\textwidth}p{0.25\textwidth}p{0.25\textwidth}p{0.25\textwidth}}
\toprule[2pt]
\textbf{Grade} & \textbf{Pronunciation} & \textbf{Intonation} & \textbf{Fluency} \\
\midrule
\endfirsthead
\multicolumn{6}{c}%
{{\bfseries \tablename\ \thetable{} -- continued from previous page}} \\
\toprule[2pt]
\textbf{Grade} & \textbf{Pronunciation} & \textbf{Intonation} & \textbf{Fluency} \\
\midrule
\endhead
\midrule \multicolumn{6}{r}{{Continued on next page}} \\ \midrule
\endfoot
\bottomrule[2pt]
\endlastfoot

Grade 5 & Pronunciation is easy to understand; with no obvious accent; individual sounds are clear. & Native-like rhythm and intonation; perfect fluency. & Almost no repetition or self-correction; perfect fluency. \\\\
Grade 4 & Pronunciation can rather easy be understood; with very few accent. & Good control of rhythm; flexible in using intonation. & Flow of speech is effortless with little hesitation. \\\\
Grade 3 & Individual sounds are generally clear but recognisable accents induced by the mother tongue. & Rhythm and intonation are generally used appropriately; with occasional unnatural effect. & Flow of speech is generally effortless with some recognisable hesitation. \\\\
Grade 2 & Obvious inaccuracies in the pronunciation of individual sounds; poor control of rhythm. & Poor control of rhythm and intonation; show recognisable influence of the speaker's mother tongue. & Flow of speech is uneven; with noticeable self-corrections and repetitions. \\\\
Grade 1 & Many pronunciation errors; strong accent induced by the mother tongue may make understanding difficult. & Broken rhythm and unnatural intonation; with a strong effect of the influence of the mother tongue. & With noticeable self-corrections repetitions and/or unnatural hesitation and long pauses. \\\\
\end{longtable}

Please Note:

\begin{itemize}
    \item The sentences you will hear might not form coherent narratives.
    \item These sentences are recorded independently and concatenated systematically.
    \item Overlook any unnatural beginnings or endings of sentences.
    \item Disregard pauses or silences between phrases as indicators of fluency.
\end{itemize}

\end{document}